\documentclass[10.5pt,twocolumn,journal,letterpaper]{IEEEtran}
\usepackage{amsmath, amsthm, amssymb}
\usepackage{url,flushend,multirow,booktabs}
\usepackage{algorithm,algorithmic}   
\usepackage{graphicx}
\usepackage{bm}
\usepackage{cite}
\usepackage{subfigure}
\usepackage{float}
\usepackage[dvipsnames]{xcolor}  
\usepackage{multirow}
\usepackage[colorlinks,citecolor=Green,urlcolor=blue,bookmarks=false,hypertexnames=true]{hyperref}
\usepackage{colortbl}
\definecolor{maroon}{cmyk}{0.08,0.04,0.00,0.06}  

\newcommand{\eg}{e.g.}
\newcommand{\ie}{i.e.}

\newcommand{\etal}{\textit{et al.}}

\renewcommand{\algorithmicrequire}{\textbf{Input:}}   
\renewcommand{\algorithmicensure}{\textbf{Output:}}  

\DeclareFixedFont{\mf}{OT1}{ptm}{m}{n}{10pt}
\DeclareFixedFont{\mfb}{OT1}{ptm}{bx}{n}{10pt}

\begin{document}
%

\title{Adversarial Attacks on Video Object Segmentation with Hard Region Discovery}
%
%
%

\author{Ping~Li,~\IEEEmembership{Member,~IEEE}, Yu~Zhang, Li~Yuan, Jian~Zhao, Xianghua~Xu, and Xiaoqin~Zhang,~\IEEEmembership{Senior Member,~IEEE}
\thanks{P.~Li, Y.~Zhang and X.~Xu are with the School of Computer Science and Technology, Hangzhou Dianzi University, Hangzhou, China (e-mail: patriclouis.lee@gmail.com, zycs@hdu.edu.cn, xhxu@hdu.edu.cn). P.~Li	is also with Guangdong Laboratory of Artificial Intelligence and Digital Economy (SZ), Shenzhen, China.}
\thanks{L.~Yuan is with the School of Electronic and Computer Engineering, Peking University, China, and also with PengCheng Laboratory, China (e-mail:yuanli-ece@pku.edu.cn).}
\thanks{J.~Zhao is with the Institute of North Electronic Equipment, Beijing, China. (e-mail:zhaojian9014@gmail.com).}
\thanks{X.~Zhang is with the College of Computer Science and Artificial Intelligence, Wenzhou University, Wenzhou, China. (e-mail:xqzhang@wzu.edu.cn).}
}
\markboth{arXiv}
{LI \MakeLowercase{\textit{et al.}}:~Adversarial Attacks on Video Object Segmentation with Hard Region Discovery}
%

\maketitle

\begin{abstract}
  Video object segmentation has been applied to various computer vision tasks, such as video editing, autonomous driving, and human-robot interaction. However, the methods based on deep neural networks are vulnerable to adversarial examples, which are the inputs attacked by almost human-imperceptible perturbations, and the adversary (\ie, attacker) will fool the segmentation model to make incorrect pixel-level predictions. This will rise the security issues in highly-demanding tasks because small perturbations to the input video will result in potential attack risks. Though adversarial examples have been extensively used for classification, it is rarely studied in video object segmentation. Existing related methods in computer vision either require prior knowledge of categories or cannot be directly applied due to the special design for certain tasks, failing to consider the pixel-wise region attack. Hence, this work develops an object-agnostic adversary that has adversarial impacts on VOS by first-frame attacking via hard region discovery. Particularly, the gradients from the segmentation model are exploited to discover the easily confused region, in which it is difficult to identify the pixel-wise objects from the background in a frame. This provides a hardness map that helps to generate perturbations with a stronger adversarial power for attacking the first frame. Empirical studies on three benchmarks indicate that our attacker significantly degrades the performance of several state-of-the-art video object segmentation models.
\end{abstract}


 \ifCLASSOPTIONpeerreview
 \begin{center} \bfseries EDICS Category: 3-BBND \end{center}
 \fi

\IEEEpeerreviewmaketitle

\section{Introduction}
\label{sec1:intro}

\IEEEPARstart{D}{riven} by the increasing demand of video editing \cite{zhang-acmmm2021-tcvom} and autonomous driving \cite{zhang-cvpr2016-autodrive}, Video Object Segmentation (VOS)  \cite{fan-tcsvt2022-semivos, xi-tcsvt2022-imcnet, sun-tcsvt2023-vos, xi-tcsvt2023-ouvos} has attracted lots of interest in both academia and industry. Essentially, VOS aims to separate the foreground (\ie, objects) and the background pixels in all video frames. When the target objects are specified in the first frame during inference, the goal of the segmentation model is to estimate the object masks in all remaining frames. Recently, great efforts have been made to investigate VOS models using deep neural networks, which are vulnerable to adversarial examples \cite{madry-iclr2018-pgd, tu-tist2020-lgidr}, \ie, the inputs are almost indistinguishable from natural data, easily leading to incorrect predictions. This rises the potential attack risks of VOS models, and the security danger is dramatically increased when these models are deployed in a highly-demanding environment.  

To this end, adversarial examples have been extensively investigated in computer vision tasks, such as image classification \cite{dong-cvpr2019-ti-fgsm, goodfellow-iclr2015-fgsm}, video classification \cite{li-nips2021-gt, pony-cvpr2021-oaafa}, object detection \cite{jia-iclr2020-mot}, object tracking \cite{chen-cvpr2020-osaa, jia-cvpr2021-iou-attack, jiang-arxiv2021-anti-uav}, and person re-identification \cite{bai-pami2021-ama}. However, few studies have explored the influences of adversarial attacks on video object segmentation. Thus, this work develops a novel adversarial attack method by discovering hard regions of frames, and shows that state-of-the-art (SOTA) VOS models are easily attacked by adversarial examples generated by simply adding some perturbations to the first frame of the video. Usually, the attack effect is evaluated in terms of the segmentation performance degradation. 

Though existing adversarial attacks in vision tasks shed some light on VOS models, they are still inappropriate for this scenario due to two-fold reasons. First, video classification and object detection both need prior knowledge about categories, and the model can be attacked by maximizing the class probability with smaller confidence, \eg, fooling the model to make an incorrect prediction with the less-possible class from multiple candidates. This is not applicable to VOS tasks whose frame pixels are either foreground or background, and there is only one candidate class to select, thus making the adversarial attack much more difficult. Second, the attacks on object tracking \cite{chen-cvpr2020-osaa} without known categories are tailored for producing wrong bounding boxes, while the attack on person re-identification \cite{bai-pami2021-ama} fools a discriminant distance metric function. Also, they cannot be directly applied to video object segmentation. 

\begin{figure*}[!t]
	\centering
	\includegraphics[width=1\linewidth]{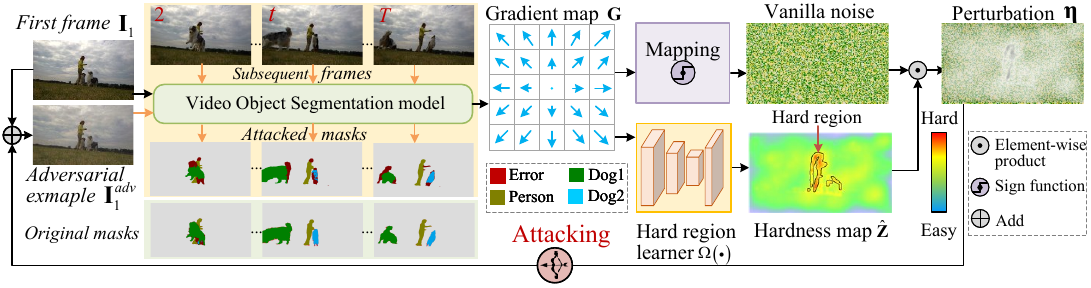}
	\caption{The entire framework of adversarial region attack on the VOS model. It contains two data flows: one is the adversarial example generation flow with hard region discovery using the gradient map derived from the first frame indicated by the \textbf{black} arrows, and the other is the adversarial attack on the VOS inference of the subsequent frames indicated by the \textbf{\textcolor{yellow}{yellow}} arrows. Compared to original masks without attacks, the attacked masks have large error proportions in \textcolor{red}{red}.}
	\label{fig:motivation}
\end{figure*}

Therefore, this paper develops an adversarial attack method for VOS (See Fig.~\ref{fig:motivation}) and concentrates on the semi-supervised setting \cite{caelles-cvpr2017-osvos, oh-iccv2019-stm}, where the ground-truth mask of the target object is given in the first frame during inference. We consider the semi-supervised VOS as it is the most widely explored setting in VOS with very cheap annotation cost only on the first frame, and it gains more popularity in practice compared to unsupervised ones (zero-shot VOS \cite{zhou-tip2020-matnet}). Naturally, this work chooses to attack only the first frame by slightly perturbing its pixel values, thus indirectly attacking the subsequent frames. Particularly, the first frame is fed to a well-trained VOS model to generate the gradient map for obtaining perturbation, which is used for attacking the first frame to generate adversarial examples. Then, the adversarial example is fed into the VOS model to fool the inference of subsequent frames, and thus more incorrect estimations are made compared to those without the attack. Here, a question may raise as to whether some pixel areas of the frame should be emphasized more for producing a stronger adversarial example. We argue that the foreground and the background are easily confused in the emphasized pixel area, which is regarded as a \emph{hard region}. Motivated by this, a hard region learner (\eg, ResNet \cite{he-cvpr2016-resnet}), is placed after the gradient map, to output a hardness map whose entries indicate whether the pixel hardness is high or low. The hardness map is involved in element-wise production with the vanilla noise map derived from mapping the gradients to a set space via a sign function, which results in stronger perturbations. Note that there may be multiple hard regions in one frame according to the emerging objects in a video. Simply, the entire attacking framework proposed in this paper is called the \emph{Adversarial Region Attack} (ARA) method. 

The objective function includes the Cross-Entropy (CE) loss of the VOS model and the $\ell_2$-norm hardness loss of the hard region learner. The VOS model produces the loss value map, whose entries indicate whether the pixels of the first video frame are difficult to discriminate from the foreground and background. Those pixels with high loss values constitute the hard region. Then, the loss value map is binarized into a hardness pseudo-label map, which provides some supervised knowledge for optimizing the hardness loss function. Hence, the hardness map reflects whether the pixels belong to the hard region or not. 

The main contributions of this paper are summarized below:
\begin{itemize}
  \item To the best of our knowledge, this paper is the first to study adversarial attacks against VOS models and propose a class-agnostic Adversarial Region Attack method to fool the model to make incorrect predictions, by generating small perturbations only on the first frame.

  \item A newly developed $\ell_2$-norm based hardness loss function is minimized to obtain the hardness scores of pixels with large confidence, under the guidance of a hardness pseudo-label map.

  \item Our attacker is evaluated on three VOS benchmarks, including DAVIS2016 \cite{perazzi-cvpr2016-davis2016}, DAVIS2017 \cite{pont-arxiv2017-davis17}, and YouTube-VOS \cite{xu-arxiv2018-ytbvos}. Besides the white-box attack, both the black-box attack and the defense are investigated. The experimental results indicate that our attacker significantly degrades the segmentation performance of several SOTA VOS methods and exhibits a stronger attack power compared to a few adaptive alternatives. Meanwhile, the defense performance of our model has been justified.

\end{itemize}

%

\section{Related Work}
\label{related}

\subsection{Semi-supervised Video Object Segmentation}
Semi-supervised Video Object Segmentation (SVOS) \cite{zhu-tcsvt2022-separable} aims to distinguish the pixel-wise object region in a video where the object is specified by the annotation of the first frame. Sometimes, it is also called one-shot VOS \cite{li-tip2022-oneshotvos}. Current SVOS approaches can be mainly divided into two categories: 1) \emph{online learning}; 2) \emph{offline learning}. The \emph{online learning} methods \cite{caelles-cvpr2017-osvos, robinson-cvpr2020-frtm, bhat-eccv2020-lwl, park-cvpr2021-reuse} update and optimize model parameters by the historical prediction mask or ground-truth mask to obtain robust appearance representations during inference. However, they require video frames to optimize model parameters during inference. When video frames are attacked by perturbations, it has adverse impacts on the model optimization, thus degrading the segmentation performance. 

Different from the online methods requiring model training during inference, the offline methods infer segmentation directly using the trained model, and they can be further divided into \emph{propagation} methods and \emph{spatio-temporal matching} methods. The propagation methods \cite{khoreva-ijcv2019-lucid, li-eccv2018-dyenet, perazzi-cvpr2017-masktrack} propagate the mask of the first frame to other frames sequentially, but they exploit the temporal consistency of nearby frames and easily suffer from drastic object deformations in long videos. By contrast, the spatio-temporal matching methods \cite{cheng-nips2021-stcn, oh-iccv2019-stm, seong-eccv2020-kmn, seong-iccv2021-hmmn} achieve better segmentation effects by using the memory network to reveal the spatio-temporal relations of the historical frames and the current frames. Thus, this paper takes the spatio-temporal matching method as the target VOS model.

\subsection{Adversarial Attacks}
Adversarial attacks are implemented by adversarial examples \cite{goodfellow-iclr2015-fgsm, szegedy-iclr2014-ipnn}, \ie, applying small but almost human-imperceptible perturbations to clean data samples. When the sample is an image, its pixels are either partially or fully perturbed, and full perturbation is considered in this paper. Then, the adversarial example is used to fool a model to make incorrect estimations, which is called an \emph{attack}. Generally, adversarial attacks can be divided into \emph{white-box} and \emph{black-box} attacks. The white-box attack \cite{goodfellow-iclr2015-fgsm, madry-iclr2018-pgd} exploits the model knowledge, including the structure, the parameters, and the trainable weights used for computing the gradients to generate the model-aware perturbations. By contrast, the black-box attack \cite{li-cpvr2021-qair} has very limited or no knowledge about the model, so it yields the model-agnostic perturbations. Our work mainly focuses on while-box attacks, and provides the empirical studies on black-box attacks.

\subsection{Adversarial Attacks in Computer Vision Tasks}
\textbf{Image Classification}. In computer vision, Szegedy \etal~\cite{szegedy-iclr2014-ipnn} first discovered that images with tiny perturbations can be used as adversarial examples to deceive image classifiers for making wrong predictions. Then, Goodfellow \etal~\cite{goodfellow-iclr2015-fgsm} pointed out that due to the naturally linear characteristics, deep neural networks are susceptible to the deception of adversarial examples. The image gradients back-propagated by a model can be used to generate perturbations to deceive the classification network. Besides, Kurakin \etal~\cite{kurakin-iclr2017-bim} iteratively updated the adversarial examples by gradually adding perturbations to the source image. Moreover, Madry \etal~\cite{madry-iclr2018-pgd} found that adding random perturbations to the source image at the initial iteration can increase the attack ability of adversarial examples. Additionally, most image classification networks adopt convolution operations, and the pixel-level features are easily affected by its adjacent regions. Therefore, Gao \etal~\cite{gao-eccv2020-pi-fgsm} adjusted the image gradients and assigned those exceeding a threshold to the neighbors of each pixel, making adversarial examples more robust. Furthermore, Dong \etal~\cite{dong-cvpr2019-ti-fgsm} proposed a translation-invariant attack method, which equips the adversarial example with more transferability by employing image gradients on translated images. 

\textbf{Image Segmentation}. Adversarial attacks \cite{arnab-pami2020-seg-attack,xie-iccv2017-dag,gu-eccv2022-segpgd} on semantic segmentation \cite{chen-eccv2018-deeplab, li-arxiv2023-trikd} are often adapted from the attack methods on image classification. For example, Arnab \etal~\cite{arnab-pami2020-seg-attack} found that the segmentation model using a deep neural network is vulnerable to adversarial examples; Xie \etal~\cite{xie-iccv2017-dag} investigated the influence of adversarial examples on object detection and semantic segmentation models simultaneously by a high-transferability attack method; Gu~\etal~\cite{gu-eccv2022-segpgd} develop a model that creates more effective adversarial examples than PGD \cite{madry-iclr2018-pgd} under the same number of attack iterations.  

\textbf{Video Classification}. For this task, Pony \etal~\cite{pony-cvpr2021-oaafa} proposed a flickering temporal perturbation to deceive the classifier to generate wrong predictions. Existing anti-attack methods for classifiers usually rely on maximizing the probability of misclassification to obtain the gradient and then convert the gradient into perturbation. To capture more effective gradients, Li \etal~\cite{li-nips2021-gt} searched for better gradient update directions through the geometric transformation of the input frames, thus generating the desired deviations for improving the attack power. Additionally, Hwang \etal~\cite{hwang-iccv2021-jom} focused on the structural vulnerability of action recognition, \ie, the influences of modeling temporal information in deep models.

\textbf{Object Tracking}. Unlike classification, object tracking \cite{tao-cvpr2016-sis, li-cvpr2019-siamrpn++,  wang-cvpr2019-siamseg} aims to capture the trajectory of the moving object in videos and produce a series of object bounding boxes. To this end, several attempts \cite{chen-cvpr2020-osaa, guo-eccv2020-spark, yan-cvpr2020-csa,jia-iclr2020-mot, jia-cvpr2021-iou-attack} have been made on the adversarial attacks for object tracking. For instance, Guo \etal~\cite{guo-eccv2020-spark} proposed an online and incremental sparse perturbation generation scheme in the spatial domain to ensure attack efficiency; Chen \etal~\cite{chen-cvpr2020-osaa} added perturbations only to the object area of the first frame. For multi-object tracking, Jia \etal~\cite{jia-iclr2020-mot} introduced a tracking error reduction process to attack the object detection and tracking model at the same time, causing the tracker to lose the object. Additionally, Jia \etal~\cite{jia-cvpr2021-iou-attack} attempted to generate more effective perturbations by optimizing the IoU (Intersection over Union) scores of the current and previous frames.

However, existing adversarial attack methods in the computer vision field are specially designed for certain tasks or models, so it is difficult to transfer them to the VOS model. For example, an adversarial attack approach for object tracking designs a regression loss against the object bounding box in the tracker to generate perturbations. Such a regression loss fails to handle the pixel-level classification task like VOS. Meanwhile, the video classification model is designed to assign the video with one label from predefined categories, but VOS is a class-agnostic task. Besides, the attack methods against the image classifier or semantic segmentation model aim to deceive the target model from one semantic class to another in predefined categories, which does not hold for VOS. Therefore, it is desirable to develop an adversarial attack method specially for the VOS task.

\section{Method}
\label{method}

\subsection{Problem Definition}
Our attacker concentrates on the most popular semi-supervised VOS models during inference. Formally, the already-trained VOS model is defined as $\Phi(\cdot)$ with the parameters $\theta$. Given a video $\mathcal{V}=\{ \mathbf{I}_t\in \mathbb{R}^{H\times W\times 3}|t=1,\ldots,T \}$ and the ground-truth mask  $\mathbf{Y}_1\in \mathbb{R}^{H\times W}$ of the first frame $\mathbf{I}_1$, the goal of the segmentation model is to produce the prediction mask $\mathbf{\hat{Y}}_t \in \mathbb{R}^{H\times W}$ of all remaining frames. Here, $H$ is the frame height, $W$ is the frame width, $t$ is the frame index, and $T$ denotes the total number of frames. Each foreground pixel of the mask is set to 1, while the background pixel is set to 0. If there are multiple objects, each object corresponds to a binary classification problem. Thus, VOS is an object class-agnostic task. In this setting, this work manipulates only the first frame by adding a small human-imperceptible perturbation $\boldsymbol{\eta}\in\mathbb{R}^{H\times W\times 3}$ to generate the adversarial example, \ie, $\mathbf{I}_{1}^{adv}= \mathbf{I}_1 + \boldsymbol{\eta}$, which is used to mislead the model to degrade the segmentation performance on the subsequent frames. Thus, the inference process produces the attacked mask sequence $\hat{\mathcal{Y}}^{adv} = \{\mathbf{\hat{Y}}_t^{adv}\}_{t=2}^T$, whose entry is represented as
\begin{equation}
	\mathbf{\hat{Y}}_t^{adv} = \Phi( \{ \mathbf{I}_{1}^{adv}, \mathbf{I}_{2}, \ldots, \mathbf{I}_{t} \}, \mathbf{Y}_1; \theta) \in \mathbb{R}^{H\times W}.
\end{equation}
To generate the perturbation $\boldsymbol{\eta}$, this paper follow the gradient-based adversary, \ie, the fast gradient sign method \cite{goodfellow-iclr2015-fgsm}, which linearizes the cost function $\mathcal{L}_1(\cdot)$ around the current value of $\theta$  and utilizes the back-propagation gradients to generate the max-norm constrained perturbation
\begin{equation}
	\label{eq:fgsm}
	\boldsymbol{\eta}_t = \epsilon \cdot 
	\text{sign}[\nabla_{\mathbf{I}_t} \mathcal{L}_1(\Phi(\mathbf{I}_t; \theta), \mathbf{Y}_t)],
\end{equation}
where $\mathcal{L}_1(\cdot)$ is the CE loss, $\nabla_{\mathbf{I}_t}$ denotes the gradient with regard to the $t$-th frame $\mathbf{I}_t$ obtained by back-propagation according to $\mathcal{L}_1(\cdot)$, the constant $\epsilon>0$ is the upper-bound perturbation value and is set to $8/255$, and $\text{sign}[\cdot]$ denotes a sign function that maps all input gradient elements to a discrete set $\{-1, 0, 1\}$. Here, $(-\epsilon, +\epsilon)$ forms a ball of the infinite norm.

\subsection{Overview of VOS Model}
In this paper, the spatio-temporal matching-based VOS model is adopted as the target model for attacking, and the early work Space Time Memory (STM) networks  \cite{oh-iccv2019-stm} of this type is taken as an example. The target model processes from the second frame in the video sequence using the ground-truth mask $\mathbf{Y}_1$ of the first frame, resulting in the prediction masks $\{ \mathbf{\hat{Y}}_2,\ldots,\mathbf{\hat{Y}}_{t-1} \}\in \mathbb{R}^{H\times W}$. When processing the current frame $\mathbf{I}_t$, the past frames with predicted object masks are used to establish a pair-wise memory set, \ie,  $\mathcal{M}=\{(\mathbf{I}_1,  \mathbf{Y}_1),  (\mathbf{I}_2,  \mathbf{\hat{Y}}_2), \ldots, (\mathbf{I}_{t-1}, \mathbf{\hat{Y}}_{t-1}) \}$. The memory set provides long-range spatio-temporal visual semantics, which helps to distinguish the object from the background in a frame. 

Meanwhile, the frame-mask pairs in the memory set are fed into one ResNet \cite {he-cvpr2016-resnet} encoder to produce the past frame feature subspace, and the current frame is fed into the other encoder to produce the current frame feature subspace. Then, spatio-temporal matching is performed in the feature subspace, \ie, the pixel class of the current frame is inferred from that in past frames (the elements of the predicted mask reflect the pixel-wise binary classes) according to the semantic similarity, generating a class feature map of the current frame. Then, this class feature map is exploited to produce the mask of the current frame by a softmax function after several convolution layers with bi-linear interpolation. Thus, the mask sequence is produced by $\mathbf{\hat{Y}}_t= \Phi(\mathcal{M}, \mathbf{I}_t; \theta) \in \mathbb{R}^{H\times W}$.

\subsection{Adversarial Region Attack}
To attack the VOS model, the perturbation (\ie, vanilla noise) in Eq.~(\ref{eq:fgsm}) is weak, as VOS is actually a pixel-wise binary classification problem. The pixel class label is either foreground or background, and the object class remains unknown. Such a binary problem increases the attack difficulty. This is because it is easy to deceive the model for randomly picking one wrong class from several candidates in a multi-class problem, but it becomes more difficult when there is only one candidate. For example, given a \emph{ten}-class problem, the probability of a successful attack is $0.9$ in theory, which decreases to 0.5 for binary classification. Inspired by this, our work develops an ARA method to strengthen the attack power of adversarial examples, and the whole framework is drawn in Fig.~\ref{fig:motivation}.

In practice, segmentation errors usually occur in the pixel area where the foreground and the background are easily confused, \eg, similar appearance, vague boundary, and salient background (like large trees). Thus, the possibly-confused pixel area is the hard region that requires more emphasis, and it is often vulnerable to perturbations. That is, perturbing the hard region of the frame can produce a stronger adversarial example. To this end, this paper designs a Hard Region Learner (HRL) to derive a hardness map, which reveals how difficult each pixel is to be correctly segmented. The hardness map and the vanilla noise in Eq.~(\ref{eq:fgsm}) are incorporated by the element-wise product operation to generate a stronger perturbation. The details are described below.

Our attacker adopts the white-box attack and exploits the image gradients obtained from the back-propagation in the model. Both the vanilla noise map and the hardness map are learned from the gradient map of the first frame $\mathbf{I}_1$, and the gradient map $\mathbf{G}$ is obtained by
\begin{equation}
	\label{eq:gradient}
	\mathbf{G}= \nabla_{\mathbf{I}_1}\mathcal{L}_1({\Phi(\mathcal{M}^\prime, \mathbf{I}_1; \theta), \mathbf{Y}_1)} \in \mathbb{R}^{H\times W\times 3},
\end{equation}
where the memory subset $\mathcal{M}^\prime$ contains two nearby frames of the first frame and their prediction masks, \ie, the second and the third frame-mask pairs. Here, using two nearby frames follows the model training process \cite{oh-iccv2019-stm}, which stacks three frames in the GPU memory for efficiency. If more powerful GPUs are available, more frames can be considered. The $\mathcal{L}_1(\cdot)$ represents the segmentation loss between the prediction mask $\hat{\mathbf{Y}}_1 = \Phi(\mathcal{M}^\prime, \mathbf{I}_1; \theta)$ and the ground-truth mask $\mathbf{Y}_1$ of the first frame. The gradient values are generally smaller than the original image pixel values, and the layer normalization \cite{ba-arxiv2016-layernorm} strategy is adopted to normalize the gradient values along each channel.

By default, the vanilla noise map is denoted as $\boldsymbol{\eta}_1 \in \mathbb{R}^{H\times W\times 3}$, which is calculated by Eq.~(\ref{eq:fgsm}). To capture the hardness map, this paper uses convolution neural networks to construct the HRL $\Omega(\cdot)$ and adopts the light ResNet18 \cite{he-cvpr2016-resnet} as the backbone.The ResNet involves four stages and each stage has a resolution downsampling ratio, \ie, $\{1/4, 1/8, 1/16, 1/32\}$, for the learned feature map. To keep a large spatial resolution of the feature map, only the former two stages, followed by one convolution layer with a kernel size of $3 \times 3$, are employed to discover the hard region from the frame. Then, the bilinear upsampling $\text{Upsample}(\cdot)$ and the sigmoid function $\sigma(\cdot)$ are performed on the obtained feature map to obtain the hardness map, \ie,
\begin{equation}
	\label{eq:hardnessmap}
	\hat{\mathbf{Z}} = \sigma(\text{Upsample}(f( \mathbf{G})) ) = \Omega(\mathbf{G}) \in \mathbb{R}^{H\times W},
\end{equation}
where $f(\cdot)$ represents the modified ResNet module, and the bilinear upsampling is employed to scale the feature map up to the same size as the input frame. The scores in the hardness map fall between 0 and 1. The higher the hardness score, the more difficult the pixel is for segmentation.

\textbf{Optimization of the HRL}. To solve the problem that the ground-truth hardness map is unavailable for optimizing the hardness loss function, this paper introduces the hardness pseudo-label map as the supervisor to guide the hard region learning. Particularly, the segmentation loss $\mathcal{L}_1(\cdot)$ calculated in the already trained VOS model is used to generate the hardness pseudo-label map. The rationale is that the segmentation loss of a well-trained model is minimized during training, and a large proportion of the pixels are expected to produce small loss values while only a small set of pixels have large values. Generally, a large loss value indicates that the corresponding pixel is difficult to separate from the foreground and background. And the CE loss is adopted as the segmentation loss.

The pixel class labels are the entries of the ground-truth mask $\mathbf{Y}\in \{0,1\}^{H\times W}$, which is flattened into a long vector $\mathbf{y} = [y_i]_{i=1}^n \in \{0,1\}^n$, where $n = H\times W$, and the index $i$ specifies the spatial position of each pixel in the mask. Similarly, the prediction mask $\hat{\mathbf{Y}}\in \mathbb{R}^{H\times W}$ is flattened into a long vector $\hat{\mathbf{y}} = [\hat{y}_i]_{i=1}^n \in \mathbb{R}^n$, whose entries denote the probability of the pixel belonging to the object class. The CE loss is defined as $-\mathbf{y}\log \hat{\mathbf{y}}$, and a thresholding strategy is adopted to generate the hardness pseudo label vector $\mathbf{z}$, whose entries are obtained by
\begin{equation}
	\label{eq:pseudo_label}
	z_i = \begin{cases}
		1,  & - {y_i}\log(\hat{y}_i) > \alpha, \\
		0, & \text{others},
	\end{cases}
\end{equation}
where $\alpha>0$ is an empirical threshold, $\log(\cdot)$ denotes the natural logarithm, and $\mathbf{z} = [z_i]_{i=1}^n \in \{0,1\}^n$. Then, the obtained pseudo-label vector is reshaped to the hardness pseudo-label map $\mathbf{Z}\in \{0, 1 \}^{H\times W}$ of the first frame. The pixels with loss values larger than the threshold indicate they are difficult to segment, and those with loss values smaller than the threshold are relatively easier to segment.

After the hardness pseudo-label map $\mathbf{Z}$ and the hardness map $\hat{\mathbf{Z}}$ are obtained, the objective function of the HRL is defined in a vector form as 
\begin{equation}
	\label{eq:hardness_loss}
	\mathcal{L}_{2} = \frac{1}{n} \|\mathbf{z} - \hat{\mathbf{z}}\|_2^2,
\end{equation}
where $\hat{\mathbf{z}}\in \mathbb{R}^n$ is a flattened vector of the matrix $\hat{\mathbf{Z}}$, and the operator $\|\cdot\|_2$ denotes the $\ell_2$ norm and is a $\ell_2$-loss. In this way, the optimal hardness map can be obtained by minimizing the error between the two hardness maps, \ie, $\mathbf{E}= \mathbf{Z} - \hat{\mathbf{Z}}$.

\subsection{Working Mechanism of Our Attacker}

During the VOS inference, only the first frame $\mathbf{I}_1$ is attacked, and its adversarial example $\mathbf{I}_1^{adv}$ is generated by imposing the perturbation $\boldsymbol{\eta}^\prime$. This perturbation considers the hard pixel areas discovered by the HRL with the vanilla noise in Eq.~(\ref{eq:fgsm}). Meanwhile, an iterative scheme is adopted to generate a strong adversarial example.

In the initial period, the perturbation is the random noise $\boldsymbol{\eta}_0^\prime \in [-\epsilon,\epsilon]^{H\times W \times 3}$, which is added to the first frame for the next iteration. The maximum iteration number is $K$, which is set to 10 for efficiency. If the perturbation in the $r$-th iteration is $\boldsymbol{\eta}_r^\prime$, then the corresponding adversarial example is represented as:
\begin{equation}
	\label{eq:adverseexample_iter}
	\mathbf{I}_{1,r}^{adv} = \text{Clip}_{\mathbf{I}_{1}, \epsilon}( \mathbf{I}_{1,r-1} + \boldsymbol{\eta}_{r-1}^\prime),
\end{equation}
where $\text {Clip}_{\mathbf{I}_1, \epsilon} (\cdot) $ squeezes the numerical range of the input frame to a range $[(\min(\mathbf{I} _1) - \epsilon, \max(\mathbf{I}_1) + \epsilon]$. Here, $\min(\cdot)$ and $\max(\cdot)$ are the minimum and the maximum function of pixel values, respectively. 

\textbf{Perturbation Update}. To update the perturbation, the gradient map $\mathbf{G}_{r}$ of the first frame like Eq.~(\ref{eq:gradient}) is first calculated according to 
\begin{equation}
	\label{eq:gradient_iter}
	\mathbf{G}_{r} = 
	\nabla_{\mathbf{I}_{1,r}^{adv}}\mathcal{L}_1(\Phi(\mathcal{M}^\prime, \mathbf{I}_{1,r}^{adv}; \theta ),\mathbf{Y}_1 ),
\end{equation}
where the model parameters $\theta$ are fixed during back-propagation. Then, the gradient map $\mathbf{G}_{r}$ is taken as the input of Eq.~(\ref{eq:hardnessmap}) to derive the hardness map $\hat{\mathbf{Z}}_r$, which is replicated for a triple of maps to form the hardness tensor $\hat{\mathbf{Z}}_r^\prime \in \mathbb{R}^{H\times W\times 3}$. Meanwhile, the gradients are projected to a set $\{-1, 0, 1\}$ by a sign function, \ie, $\mathbf{G}_r^\prime = \text{sign}(\mathbf{G}_r)$. Finally, an element-wise product is made between the hardness tensor and the sign gradient tensor, resulting in the updated perturbation as
\begin{equation}
	\label{eq:noise_iter}
	\boldsymbol{\eta}_{r}^\prime = \beta \mathbf{\hat{Z}}'_{r} \odot \text{sign}(\mathbf{G}_{r} ),
\end{equation}
where the constant $\beta>0$ governs the numerical range of the perturbation, and $\odot$ denotes an element-wise product. As the iteration continues, the perturbation ability is strengthened, thus generating a stronger adversarial example. The final adversarial example $\mathbf{I}_{1,K}^{adv}$  is fed into the VOS model to generate a sequence of attacked masks.

Besides Eq.~(\ref{eq:noise_iter}), other perturbation generation schemes \cite{gao-eccv2020-pi-fgsm, madry-iclr2018-pgd, wang-cvpr2021-vmi-fgsm} also using the gradient-based adversarial attack can be adopted by simply adding the element-wise product of the hardness map to its vanilla noise. This enables our attacker to be generalized to more scenarios easily.

\subsection{Adversarial Region Attack (ARA) Algorithm}

Our proposed Adversarial Region Attack (ARA) method is a white-box attacker, and its primary procedure is briefly summarized in Algorithm~\ref{alg:ara-training}. The ARA attacker is developed for attacking an already well-trained VOS model $\Phi$ with the parameter $\theta$. 

Given a video sequence $ \mathcal{V}$ with $T$ frames and the first frame mask $\mathbf{Y}_1\in\mathbb{R}^{H\times W}$, our attacker obtains the gradient map $\mathbf{G}\in\mathbb{R}^{H\times W\times 3}$ of the first frame. Then, by using the gradient map, the hardness map $\mathbf{\hat{Z}}_r\in \mathbb{R}^{H\times W}$ is obtained via the HRL (ResNet). Next, the vanilla noise and the hardness map are unified, and the attacker produces an adversarial example $\mathbf{I}_{1,r+1}^{adv}$ by iteration. Finally, by replacing the first frame with the adversarial example as the input of the VOS model, our attacker can fool the model to degrade the segmentation performance.

\begin{algorithm}
	\caption{Adversarial Region Attack (ARA) for VOS.}
	\label{alg:ara-training}
	\small
	\begin{algorithmic}[1]
		\REQUIRE A video sequence $\mathcal{V}$ with $T$ frames, the first frame mask $\mathbf{Y}_1\in\mathbb{R}^{H\times W}$, the trained VOS model $\Phi$ with the parameter $\theta$, the maximum iteration $K=10$, the HRL $\Omega(\cdot)$.
		\ENSURE An adversarial example $\mathbf{I}_{1,K}^{adv}$.
		\STATE Initialize the weights of the HRL $\Omega(\cdot)$ with Kaiming initialization \cite{he-iccv2015-he-init};
		\STATE Fix the model parameter $\theta$ of the VOS model $\Phi$;
		\STATE Add randomly initialized perturbations to the first frame to obtain an adversarial example $\mathbf{I}_{1,0}^{adv}$;
		\FOR{$r=1,\ldots,K$}
		\STATE Feed the adversarial example $\mathbf{I}_{1,r}^{adv}$ and video frames to the VOS model to calculate the segmentation loss;
		\STATE Obtain the gradient map $\mathbf{G}\in\mathbb{R}^{H\times W\times 3}$ by optimizing the segmentation loss;
		\STATE Feed the gradient map $\mathbf{G}$ to the HRL $\Omega(\cdot)$ to obtain the hardness map $\mathbf{\hat{Z}}_r\in \mathbb{R}^{H\times W}$;
		\STATE Obtain the hardness pseudo-label map $\mathbf{Z}\in \mathbb{R}^{H\times W}$ by using the segmentation loss with a thresholding strategy;
		\STATE Optimize the hardness loss for training the HRL $\Omega(\cdot)$;
		\STATE Generate the current perturbation $\boldsymbol{\eta}'_{r} \in [ -\epsilon, \epsilon ]^{H\times W \times 3}$ with the gradient map $\mathbf{G}$ and the hardness map $\mathbf{\hat{Z}}_r$;
		\STATE Update the adversarial example $\mathbf{I}_{1,r+1}^{adv}$ with the current adversarial example $\mathbf{I}_{1,r}^{adv}$ and perturbation $\boldsymbol{\eta}'_{r} $.
		\ENDFOR
	\end{algorithmic}
\end{algorithm}

Besides the white-box ARA attacker, this paper provides its black-box version since the model structure, parameters, and gradients are unknown to users in many situations. Most of the procedures are the same as those in Algorithm~\ref{alg:ara-training} except for the perturbation update. For the black-box attack, the initial perturbation is also the random noise added to the first frame and is updated in iteration without gradients, but the perturbation update is simple: 
\begin{equation}
	\label{eq:noise_iter_black}
	\boldsymbol{\eta}_{r}^\prime = \beta \mathbf{\hat{Z}}'_{r},
\end{equation}
where the constant $\beta>0$ governs the numerical range of the perturbation.

In addition, this paper explores the adversarial training of the white-box ARA attacker to investigate its defense performance. For the pre-trained VOS model $\Phi(\cdot)$ with the parameter $\theta$, the perturbation derived from the ARA attacker is added to the first frame to obtain the attacked training set. Then, the model $\Phi(\cdot)$ is trained by using the attacked training set, and the segmentation performance of the updated VOS model $\Phi^\prime(\cdot)$ is investigated.

\section{Experiments}
\label{test}

All experiments were performed on a server equipped with two TITAN RTX graphics cards. The codes are compiled for PyTorch 1.10, Python 3.9, and CUDA 11.0.

\subsection{Datasets}
\textbf{DAVIS2016} \cite{perazzi-cvpr2016-davis2016}\footnote{https://davischallenge.org/davis2016/code.html} contains a total of 50 video sequences, and there are 3,455 video frames with ground-truth (GT) annotations. Each video sequence contains only a single object, and there are 50 objects in total. The video contents mainly include animals, sports, vehicles, etc. Here, 30 video sequences are used for training and 20 video sequences are used for validation.

\textbf{DAVIS2017} \cite{pont-arxiv2017-davis17}\footnote{https://davischallenge.org/davis2017/code.html} is expanded on DAVIS2016 by increasing the number of videos to 150, and there are 10,459 video frames with GT annotations. Meanwhile, annotations for multiple objects are added, and there are 376 objects in total. The dataset is split into four subsets, namely, training set, validation set, test-dev set, and test-challenge set. Among them, the training set contains 60 videos, while the validation set contains 30 videos, and GT mask annotations are provided.

\textbf{YouTube-VOS} \cite{xu-arxiv2018-ytbvos}\footnote{https://competitions.codalab.org/competitions/20127} has three subsets, namely, training set, validation set, and test set. Among them, the training set contains 3,471 videos, and there are 65 object categories, with a total of 6,459 object instances; the validation set contains 507 videos with 1,063 object instances, and 65 of its object categories also appear in the training set, while 26 categories do not; the test set contains 541 videos with 1,092 object instances, and 65 object categories also appear in the training set, while 29 categories do not. Note that each video in the validation set only provides the GT mask of the first frame, and the final evaluation results need to be uploaded to the official server. 

\textbf{A2D Sentences} \cite{gavrilyuk-cvpr2018-aavs}\footnote{https://kgavrilyuk.github.io/publication/actor\_action/} is extended from the Actor-Action Dataset \cite{xu-cvpr2015-a2d} by adding textual descriptions for each video. It contains 3782 videos annotated with 8 action classes performed by 7 actor classes and  6,655 sentences. For each video, there are 3 to 5 frames annotated with pixel-wise segmentation masks. The dataset is split into a training set and a test set with 3,036 and 746 videos, respectively. 

Among them, and the attack performance for semi-supervised VOS models on the former three datasets is examined, which is the focus of this work. Without loss of generality, all experimental results are reported on the validation set except for A2D Sentences. Additionally, the performance of our attacker for several unsupervised VOS models and referring VOS models is investigated by using the DAVIS2016 \cite{perazzi-cvpr2016-davis2016} validation set and the A2D Sentences \cite{gavrilyuk-cvpr2018-aavs} test set, respectively.

\subsection{Evaluation Metrics}
Following previous works \cite{oh-iccv2019-stm}\cite{seong-iccv2021-hmmn}, the same evaluation metrics on the benchmarks are adopted in this paper. For DAVIS2016 and DAVIS2017 datasets, \textit{region similarity} $\mathcal{J}$ and \textit{contour accuracy} $\mathcal{F}$ \cite{perazzi-cvpr2016-davis2016} are used, where the former measures the IoU ratio between the prediction mask and the ground-truth mask, and the latter measures the F1 score of the predicted and ground-truth masks at the object contour pixels. Overall, $\mathcal{J}$\&$\mathcal{F}$ represents the mean of region similarity and contour accuracy, which evaluates the overall segmentation performance of the VOS model.

For YouTube-VOS \cite{xu-arxiv2018-ytbvos}, this paper uses the same $\mathcal{J}$ and $\mathcal{F}$ provided by the official server \footnote{https://youtube-vos.org/dataset/vos/}. Since the semantic categories of some objects in the validation set do not appear in the training set, the validation set is further divided into two subsets, \ie, seen and unseen sets, where the seen subset contains the videos with seen categories in the training set, and the unseen subset contains the videos with unseen categories in the training set. The average of each metric is calculated on each subset to obtain $\mathcal{J}_{seen}, \mathcal{J}_{unseen}, \mathcal{F}_{seen}$, and $\mathcal{F}_{unseen}$. The global index $\mathcal{G}$ is the mean of the above four metrics of the seen and unseen subsets.

\subsection{Experimental Setup}
\label{sec:exp_set}
The first frame of the videos in the validation set is attacked by adding almost human-imperceptible perturbations. The maximum perturbation value $\epsilon$ in Eq.~(\ref{eq:adverseexample_iter}) and the constant $\beta$ used in Eq.~(\ref{eq:noise_iter}) are both set to $8/255$, while the threshold $\alpha$ is set to $-\log(0.4)$. During the perturbation update, the maximum iteration number of the iteration-based attack is set to 10, and the Adam optimizer \cite{kingma-iclr2015-adam} is adopted to obtain the gradient map with a learning rate of 0.1 and a weight decay of 0.01.

\begin{table*}[!h]
	\centering
	\caption{The performance of semi-supervised VOS model with different attackers on DAVIS2016 \cite{perazzi-cvpr2016-davis2016} and DAVIS2017 \cite{pont-arxiv2017-davis17} in terms of $\mathcal{J}$\&$\mathcal{F}$.   
	\label{table:result-davis}}
	\small
	\setlength{\tabcolsep}{0.8mm}{
		\begin{tabular}{lcccccccccccc}
			\toprule[0.75pt]
			\multirow{3}{*}{Attacker} & \multirow{3}{*}{Venue} & \multicolumn{5}{c}{DAVIS2016~\cite{perazzi-cvpr2016-davis2016}} &  & \multicolumn{5}{c}{DAVIS2017~\cite{pont-arxiv2017-davis17}} \\ \cline{3-7} \cline{9-13} 
			&  & \scriptsize{STM\cite{oh-iccv2019-stm}} & \scriptsize{HMMN\cite{seong-iccv2021-hmmn}} & \scriptsize{STCN\cite{cheng-nips2021-stcn}} & \scriptsize{AOTL-R\cite{yang-nips2021-aot}} & \scriptsize{AOTL-S\cite{yang-nips2021-aot}} & ~& \scriptsize{STM\cite{oh-iccv2019-stm}} & \scriptsize{HMMN\cite{seong-iccv2021-hmmn}} & \scriptsize{STCN\cite{cheng-nips2021-stcn}} & \scriptsize{AOTL-R\cite{yang-nips2021-aot}} & \scriptsize{AOTL-S\cite{yang-nips2021-aot}} \\ 
			&  & \scriptsize{ICCV'19} & \scriptsize{ICCV'21}  & \scriptsize{NeurIPS'21} & \scriptsize{NeurIPS'21} & \scriptsize{NeurIPS'21} &~ & \scriptsize{ICCV'19} & \scriptsize{ICCV'21} & \scriptsize{NeurIPS'21} & \scriptsize{NeurIPS'21} & \scriptsize{NeurIPS'21}  \\ \midrule[0.5pt]
			\rowcolor{maroon} Origin & - & 89.3 & 90.8 & 91.6 & 91.7 & 92.3 &~ & 81.8 & 84.7 & 85.4 & 85.2 & 87.0 \\
			Random & - & 88.5 & 90.7 & 91.5 & 91.6 & 92.2 & ~& 81.7 & 84.0 & 85.3 & 85.1 & 86.9 \\
			\rowcolor{maroon} FGSM\cite{goodfellow-iclr2015-fgsm} & \scriptsize{ICLR'15} & 86.7 & 88.8 & 90.5 & 90.3 & 91.5 & ~& 78.5 & 82.1 & 83.9 & 85.0 & 86.5 \\
			BIM\cite{kurakin-iclr2017-bim} & \scriptsize{ICLR'17} & 86.0 & 85.6 & 87.6 & 89.5 & 90.6 &~ & 77.7 & 82.0 & 83.7 & 83.5 & 85.3 \\
			\rowcolor{maroon} PGD\cite{madry-iclr2018-pgd} & \scriptsize{ICLR'18} & 84.4 & 85.3 & 86.9 & 82.7 & 83.9 & ~& 77.2 & 81.2 & 83.3 & 80.8 & 81.5 \\
			TI\cite{dong-cvpr2019-ti-fgsm} & \scriptsize{CVPR'19} & 88.3 & 89.3 & 88.9 & 91.2 & 91.9 &~ & 78.6 & 82.1 & 84.3 & 85.0 & 86.5 \\
			\rowcolor{maroon} PI\cite{gao-eccv2020-pi-fgsm} & \scriptsize{ECCV'20} & 84.5 & 86.1 & 87.1 & 83.1 & 85.9 & ~& 77.3 & 81.3 & 83.4 & 81.0 & 81.5 \\
			VMI\cite{wang-cvpr2021-vmi-fgsm} & \scriptsize{CVPR'21} & 84.6 & 86.9 & 87.1 & 90.9 & 91.7 &~ & 77.6 & 81.7 & 83.9  & 84.2 & 85.9 \\
			\rowcolor{maroon} \scriptsize{AutoAttack\cite{croce-icml2020-autoattack}} & \scriptsize{ICML'20} &84.2&84.7 & 86.5 &82.5 &84.0 &~ &76.8   & 80.5 &83.3 & 80.5 & 81.7 \\
			SegPGD\cite{gu-eccv2022-segpgd} & \scriptsize{ECCV'22} & 84.0 & 84.4 & 86.3 &82.4 & 83.2 &~ &76.7  &80.3 &83.2  &80.1 &80.9 \\ 
			\rowcolor{maroon}  ALMA\cite{rony-cvpr2023-alma} & \scriptsize{CVPR'23}  &84.0 & 84.2 &86.1 &82.1  &83.4 &~ & 76.4&80.2 &83.0 &80.2 & 80.7 \\
			\midrule[0.5pt]
			ARA  &  Ours & \textbf{82.1} & \textbf{82.4} & \textbf{84.7} & \textbf{80.1} & \textbf{81.7} &~ & \textbf{75.0} & \textbf{78.6} & \textbf{81.3} & \textbf{78.5} & \textbf{79.5} \\ 
			\toprule[0.75pt]
		\end{tabular}
	}
\end{table*}

\subsection{Compared Methods}
\label{sec:exp_compare}
\textbf{Attackers}. To comprehensively evaluate the effect of our ARA attacker against the VOS model, several gradient-based adversarial attack methods are taken for comparison, including FGSM \cite{goodfellow-iclr2015-fgsm}, BIM (Basic Interactive Method) \cite{kurakin-iclr2017-bim}, PGD (Projected Gradient Descent) \cite{madry-iclr2018-pgd}, TI (Translation-Invariant attack Method) \cite{dong-cvpr2019-ti-fgsm}, PI (Patch-wise Iterative FGSM) \cite{gao-eccv2020-pi-fgsm}, VMI (Variance tuning Momentum Iterative FGSM) \cite{wang-cvpr2021-vmi-fgsm}, AutoAttack \cite{croce-icml2020-autoattack}, SegPGD \cite{gu-eccv2022-segpgd}, and ALMA (Augmented Lagrangian Minimal Adversarial perturbation) \cite{rony-cvpr2023-alma}. Among them, FGSM is an non-iterative attack methods. FGSM is proposed by Goodfellow \etal~\cite{goodfellow-iclr2015-fgsm}, who pointed out that the vulnerability of deep neural networks comes from its linear characteristics. Note that our method just uses the already trained VOS models, and other attackers need modification for attacking these models.

\textbf{Semi-supervised VOS models}. This paper compares the attack power of different adversarial attack methods on several spatio-temporal matching based semi-supervised video object segmentation (SVOS) models, including STM \cite{oh-iccv2019-stm}, HMMN \cite{seong-iccv2021-hmmn}, STCN \cite{cheng-nips2021-stcn}, and AOT (Associating Objects with Transformers) \cite{yang-nips2021-aot}. Note that AOT Large version with ResNet50 backbone (AOTL-R) and the Large version with Swin Transformer \cite{liu-iccv2021-swin} backbone (AOTL-S) are adopted here. 

\textbf{Unsupervised VOS models}. They segment the objects in a video without any user annotation \cite{zhuo-tip2020-uvosonline}\cite{li-arxiv2023-ftea}, and our attacker is also applied to several unsupervised VOS models, including COSNet (CO-attention Siamese Network) \cite{lu-cvpr2019-cosnet}, MATNet (Motion-Attentive Transition Network) \cite{zhou-tip2020-matnet}, and FSNet (Full-duplex Strategy Network) \cite{ji-iccv2021-fsnet}. Among them, COSNet adopts a global co-attention mechanism to capture the inherent correlation across all video frames in the video, and it only utilizes the appearance feature, thus avoiding the time-consuming optical flow extraction like MATNet and FSNet.

\textbf{Referring VOS models}. They segment the objects in a video with textual descriptions \cite{yang-tip2022-rvos}\cite{li-arxiv2023-lsta}, and the referring VOS models examined in this paper include RefVOS (Referring video object segmentation) \cite{bellver-mta2023-refvos}, MTTR (Multimodal Tracking Transformer) \cite{botach-cvpr2022-mttr}, and ReferFormer \cite{wu-cvpr2022-referformer}. Among them, RefVOS utilizes a semantic segmentation model along with a language processing model to segment the language referred target object frame by frame, but it fails to incorporate rich spatial temporal features of the video. MTTR handles both text and frames in a single transformer. It not only captures rich spatial-temporal features but also associates the language feature and video feature at both word and pixel levels. Similar to MTTR, ReferFormer is also a transformer-based approach with a better feature backbone and more complex network architecture that requires more training data, and it has three versions, \ie, ReferFormer-T/S/L, where the ``T/S/L" indicates a tiny, small, and large version of video Swin Transformer \cite{liu-cvpr2022-vswin}.

\subsection{Quantitative Results}
\subsubsection{Semi-supervised VOS Setting}

\begin{table*}[!t]
	\centering
	\caption{The performance of semi-supervised VOS model with different attackers on YouTube-VOS \cite{xu-arxiv2018-ytbvos} in terms of $\mathcal{J}$\&$\mathcal{F}$. }
	\label{table:result-youtube}
	\small
	\setlength{\tabcolsep}{1.2mm}{
		\begin{tabular}{llccccccccc}
			\toprule[0.75pt]
			\multirow{2}{*}{VOS model}  & \multirow{2}{*}{Venue} & \multirow{2}{*}{Origin} & \multirow{2}{*}{Random} & FGSM\cite{goodfellow-iclr2015-fgsm} & BIM\cite{kurakin-iclr2017-bim} & PGD\cite{madry-iclr2018-pgd} & TI\cite{dong-cvpr2019-ti-fgsm} & PI\cite{gao-eccv2020-pi-fgsm} & VMI\cite{wang-cvpr2021-vmi-fgsm} & \multirow{2}{*}{Ours} \\
			&  &  &  & \scriptsize{ICLR'15} & \scriptsize{ ICLR'17} & \scriptsize{ICLR'18} & \scriptsize{CVPR'19} & \scriptsize{ECCV'20} & \scriptsize{CVPR'21} &   \\  \midrule[0.5pt]
			\rowcolor{maroon}STM\cite{oh-iccv2019-stm}  &\scriptsize{ICCV'19} & 79.3 & 79.2 & 77.4 & 75.3 & 75.2 & 77.1 & 76.6 & 76.0 & $\textbf{72.8}$ \\
			HMMN\cite{seong-iccv2021-hmmn}  & \scriptsize{ICCV'21} & 82.5 & 82.4 & 80.6 & 79.9 & 78.6 & 80.7 & 79.1 & 79.6 & $\textbf{76.5}$ \\
			\rowcolor{maroon}STCN\cite{cheng-nips2021-stcn}   & \scriptsize{NeurIPS'21} & 84.2 & 84.2 & 82.3 & 82.0 & 81.7 & 83.3 & 82.7 & 82.9 &  $\textbf{79.6}$ \\ 
			AOTL-R\cite{yang-nips2021-aot}  & \scriptsize{NeurIPS'21} & 84.4 & 84.1  & 83.2  & 82.9 & 82.1 & 83.3 & 82.3  & 82.6  & \textbf{80.0} \\ 
			\rowcolor{maroon}AOTL-S\cite{yang-nips2021-aot}  & \scriptsize{NeurIPS'21} & 84.7 & 84.5 & 83.4 & 83.1 & 82.3 & 83.5  & 82.8 & 83.1 &  \textbf{80.3} \\ 
			\toprule[0.75pt]
		\end{tabular}
	}
\end{table*}
%

\begin{table*}[!t]
	\centering
	\caption{The performance of the unsupervised VOS model with our attacker on the DAVIS2016\cite{perazzi-cvpr2016-davis2016} val set. }
	\label{table:uvos-first-all_davis16}
	\small
	\setlength{\tabcolsep}{3.5mm}{
		\begin{tabular}{llcccccccccc}
			\toprule[0.75pt]
			\multirow{2}{*}{VOS model} & \multirow{2}{*}{Venue} & \multicolumn{1}{l}{\multirow{2}{*}{ARA}} & \multicolumn{4}{c}{First frame} &  & \multicolumn{4}{c}{All frames} \\ \cline{4-7} \cline{9-12} 
			&  & \multicolumn{1}{l}{} & $\mathcal{J}$ &$\mathcal{F}$& $\mathcal{J}$\&$\mathcal{F}$ & Gains &  & $\mathcal{J}$ & $\mathcal{F}$ & $\mathcal{J}$\&$\mathcal{F}$ & Gains \\ 
			\midrule[0.5pt]
			\rowcolor{maroon}COSNet\cite{lu-cvpr2019-cosnet} & CVPR'19 &  & 80.5 & 79.4 & 80.0 & - & & 80.5 & 79.4 & 80.0 & -  \\
			COSNet &   & \checkmark & 77.7 & 76.8 & 77.3 & -2.7 &  &    44.5 & 33.6 & 39.1 & -40.9\\
			\rowcolor{maroon}MATNet\cite{zhou-tip2020-matnet} & TIP'20 &  & 81.4 & 80.4 & 80.9 & - & &  81.4 & 80.4 & 80.9 & -  \\
			MATNet &  & \checkmark & 78.9 & 77.3 & 78.1 & -2.8 &  & 36.2 & 36.6 & 36.4 & -43.5 \\
			\rowcolor{maroon}FSNet\cite{ji-iccv2021-fsnet} & ICCV'21 &  & 82.3 & 83.3 & 82.8 & - &  &  82.3 & 83.3 & 82.8 & - \\
			FSNet &  &\checkmark  & 79.8 & 81.2 & 80.5 & -2.3 & &  47.4 & 48.9 & 48.2 & -34.6  \\ 
			\toprule[0.75pt]
		\end{tabular}
	}
\end{table*}

The experimental results on DAVIS2016 and DAVIS2017 are reported in Table~\ref{table:result-davis}, while those on YouTube-VOS are presented in Table~\ref{table:result-youtube}. In the tables, ``Origin'' refers to the VOS model without attack, and ``Random'' refers to the adversarial attack with random noise. The best attack records are highlighted in bold.

\textbf{Results on DAVIS2016}. As shown on the left of Table~\ref{table:result-davis}, the VOS models are robust to random noise on the second row. However, the segmentation performance degrades significantly when the attacker is applied to the model. Among the competitive attackers, our attacker exhibits the strongest perturbation ability, as indicated by the bottom row, \eg, it reduces the segmentation performance by 7.2\%, 8.4\%, 6.9\%, 11.6\%, and 10.6\% on STM \cite{oh-iccv2019-stm}, HMMN \cite{seong-iccv2021-hmmn}, STCN \cite{cheng-nips2021-stcn}, AOTL-R \cite{yang-nips2021-aot}, and AOTL-S \cite{yang-nips2021-aot}, respectively. The results demonstrate the superiority of our ARA method, and this is attributed to that the HRL captures a hardness map to strengthen the perturbation. Among the five VOS models, STCN is the most robust against our attacker, and AOTL-R is the most vulnerable to being seriously attacked. 

Except for our attacker, SegPGD \cite{gu-eccv2022-segpgd} and ALMA \cite{rony-cvpr2023-alma} perform better than others among the remaining attackers, \eg, on the STM model, SegPGD is better than FGSM \cite{goodfellow-iclr2015-fgsm}, BIM \cite{kurakin-iclr2017-bim}, TI \cite{dong-cvpr2019-ti-fgsm}, PI \cite{gao-eccv2020-pi-fgsm}, and VMI \cite{wang-cvpr2021-vmi-fgsm}, by 2.7\%, 2.0\%, 4.3\%, 0.5\%, and 0.6\%, respectively. Meanwhile, TI performs the worst, and the reason is that TI generates a perturbation over an ensemble of translated images to increase the transferability, but the image translation operation possibly makes the gradient deviate from that without translation. Among these attackers, FGSM, BIM, and PGD degrade the segmentation performance but with obvious differences. Taking the attack on STCN for example, FGSM causes a performance drop of 1.1\%, and its successor BIM decreases the evaluation metric to 87.6\%, \ie, a large drop of 4.0\%, verifying the effectiveness of adopting the iteration strategy of generating the adversarial example by BIM. Based on BIM, PGD adds random noise initialization and further degrades the performance to 86.9\%, which validates the necessity of noise initialization. Moreover, it can be seen that the recently proposed attackers like TI, PI, and  VMI fail to improve the attack ability. This is because they focus on improving the transferability of the attacker on different target models, resulting in inferior performance. Among the three attackers, PI shows a rather strong attack power as it applies the the patch-wise perturbation to video frame. 
	
From the table, AutoAttack performs slightly better than PGD but has weaker attack ability compared to our ARA attacker, since it is an improved PGD and adaptively changes the attack step for image classification rather than VOS task. In addition, SegPGD and ALMA have stronger attack ability compared to PGD but are inferior to ours. This might be the reason that SegPGD is likely to attack those correctly-classified pixels neglecting those challenging pixels with more uncertainty, while ALMA desires expensive costs to seek the attack with small perturbations (\eg, less than $\epsilon=8/255$). However, these attackers cannot be directly applied to VOS without modification, and they perform unsatisfactorily when they are not designed specially for the segmentation task. Instead, this paper designs an attack framework especially considering the easily-confused pixel areas by introducing the HRL, which helps to generate adversarial examples with stronger attack power.

\textbf{Results on DAVIS2017}. The right of Table~\ref{table:result-davis} shows the similar behaviors of the attackers on the VOS models. The overall segmentation performance is lower than that on DAVIS2016. This is because there are multiple objects to be segmented in the videos of DAVIS2017, which is more challenging. Our attacker ranks the first consistently on several VOS models, indicating the effectiveness of the proposed adversarial region attack. Among the VOS models, STCN is the most robust to adversarial attacks, which may because the external examples used in training improve its robustness.

\textbf{Results on YouTube-VOS}. As shown in Table~\ref{table:result-youtube}, our attacker largely outperforms other alternatives consistently on SOTA segmentation models. For instance, our method degrades the evaluation metric by 6.5\% on the STM model. The attackers on VOS models behave similarly to that for DAVIS2016. However, the overall attack performance is the lowest among the three benchmarks, this is because the target object does not always appear in the first frame of videos in YouTube-VOS.

\begin{table*}[!t]
	\centering
	\caption{The performance of the referring VOS model with our attacker on the A2D Sentences~\cite{gavrilyuk-cvpr2018-aavs} test set.}
	\label{table:rvos-all}
	\small
	\setlength{\tabcolsep}{3.5mm}{
		\begin{tabular}{lcccccccccc}   
			\toprule[0.75pt]
			\multirow{2}{*}{VOS model} & \multirow{2}{*}{Venue} & \multirow{2}{*}{ARA} & \multicolumn{5}{c}{Precision} & mAP & \multicolumn{2}{c}{IoU} \\ \cline{4-8}  \cline{10-11} 
			&  &  &  0.5 & 0.6 & 0.7 & 0.8 & 0.9 & 0.5:0.95 & Overall  & Mean  \\ \midrule[0.5pt]
			\rowcolor{maroon}RefVOS\cite{bellver-mta2023-refvos} & MTA'23 &  & 57.6 & 53.4 & 45.6 & 31.1 & ~9.2 & - & 67.3 & 49.7 \\
			RefVOS &  & \checkmark  & 28.4 & 22.2 & 14.8 & ~7.1 & ~0.6 & - & 34.0 & 26.3 \\
			\rowcolor{maroon}MTTR\cite{botach-cvpr2022-mttr} & CVPR'22   &  & 72.1 & 68.4 & 60.7 & 45.6 & 16.4 & 44.7 & 70.2 & 61.8 \\
			MTTR &  & \checkmark & 37.3 & 29.5 & 20.4 & ~9.3 & ~1.2 & ~9.6 & 35.5 & 34.4 \\
			\rowcolor{maroon}ReferFormer-T\cite{wu-cvpr2022-referformer} & CVPR'22  &  & 82.8 & 79.2 & 72.3 & 55.3 & 19.3 & 52.8 & 77.6 & 69.6 \\
			ReferFormer-T &  & \checkmark & 43.0 & 36.6 & 27.2 & 13.9 & ~2.3 & 15.1 & 39.3 & 37.3 \\
			\rowcolor{maroon}ReferFormer-S\cite{wu-cvpr2022-referformer} & CVPR'22  &  & 82.6 & 79.4 & 73.1 & 57.4 & 21.1 & 53.9 & 77.7 & 69.8 \\
			ReferFormer-S &  & \checkmark & 42.9 & 36.2 & 27.2 & 14.9 & ~2.4 & 15.7 & 40.0 & 36.6 \\
			\rowcolor{maroon}ReferFormer-L\cite{wu-cvpr2022-referformer} & CVPR'22  &  & 83.1 & 80.4 & 74.1 & 57.9 & 21.2 & 55.0 & 78.6 & 70.3 \\
			ReferFormer-L &  & \checkmark & 47.6 & 41.0 & 31.2 & 16.0 & ~2.1 & 18.8 & 43.8 & 39.6 \\
			\toprule[0.75pt]
		\end{tabular}
	}
\end{table*}

\subsubsection{Unsupervised VOS Setting}
Table~\ref{table:uvos-first-all_davis16} shows the segmentation performance of several unsupervised VOS models attacked by our attacker on the DAVIS2016 \cite{perazzi-cvpr2016-davis2016} validation set. The perturbation is added to the first frame and all frames of each video, respectively. As shown in the table, the overall performance ($\mathcal{J}$\&$\mathcal{F}$) drops by 2.7\%, 2.8\% and 2.3\%, for COSNet \cite{lu-cvpr2019-cosnet}, MATNet \cite{zhou-tip2020-matnet}, and FSNet \cite{ji-iccv2021-fsnet}, respectively. The performance drops are much smaller than those for semi-supervised VOS model with our ARA attacker, \ie, 6.9\% to 11.6\% in Table~\ref{table:result-davis}. This is because common unsupervised models mainly utilize both appearance features and motion features to capture the target object, while semi-supervised models rely on the first frame to provide the prior knowledge about the object. Thus, unsupervised models are more robust to our first-frame attacker than semi-supervised models. However, if perturbations are added to all of the video frames, the segmentation performance drops significantly by 40.9\% , 43.5\%, and 34.6\% on COSNet \cite{lu-cvpr2019-cosnet}, MATNet \cite{zhou-tip2020-matnet}, and FSNet \cite{ji-iccv2021-fsnet} respectively, in terms of $\mathcal{J}$\&$\mathcal{F}$. This demonstrates the superiority and good transferability of our ARA attacker in an unsupervised setting. 

\subsubsection{Referring VOS Setting}
Table~\ref{table:rvos-all} shows the segmentation performance of several referring VOS models attacked by our attacker on the A2D Sentences~\cite{gavrilyuk-cvpr2018-aavs} test set. This setting requires natural language to guide the segmentation, and perturbations are added to all the video frames. 

According to the table, the segmentation performance drops greatly by 23.4\%, 27.4\%, 32.3\%, 33.2\%, and 30.7\% respectively for RefVOS \cite{bellver-mta2023-refvos}, MTTR \cite{botach-cvpr2022-mttr}, and ReferFormer-T/S/L \cite{wu-cvpr2022-referformer} in terms of the mean IoU. This suggests that our ARA attacker is also strong on referring VOS models, as the perturbed video frames successfully fool the model to make incorrect predictions on a large number of pixels.

\subsection{Ablation Study}
\begin{table}[!t]
		\centering
		\caption{Different backbones of HRL. }
		\label{table:abl-backbone}
		\small
		\setlength{\tabcolsep}{0.5mm}{
			\begin{tabular}{lrrcccc}
				\toprule[0.75pt]
				Backbone & \scriptsize{GFLOPs} & \#Params & Sec & $\mathcal{J}$  & $\mathcal{F}$  & $\mathcal{J}$\&$\mathcal{F}$ \\
				\midrule[0.5pt]
				\rowcolor{maroon}VGG16\cite{simonyan-iclr2015-vgg} & 65.4 & 14.7M & 6.4 & 79.3 & 86 & 82.7 \\
				MobileNetV3\cite{howard-iccv2019-mobilenetv3} & $\textbf{1.0}$ & $\textbf{3.0M}$ & 6.0 & 79.7 & 86.4 & 83.1 \\
				\rowcolor{maroon}InceptionV3\cite{szegedy-cvpr2016-inceptionv3} & 14.7 & 21.8M & 6.4 & 80.1 & 86.8 & 83.5 \\
				DenseNet121\cite{huang-cvpr2017-densenet} & 12.3 & 7.0M & 6.6 & 80.1 & 86.6 & 83.4 \\
				\rowcolor{maroon}ResNet18\cite{he-cvpr2016-resnet} & 7.8 & 11.2M & $\textbf{5.8}$ & $\textbf{78.4}$ & $\textbf{85.5}$ & $\textbf{82.0}$ \\
				ResNet34\cite{he-cvpr2016-resnet} & 15.7 & 21.3M & 6.0 & 79.3 & 85.9 & 82.6 \\
				\rowcolor{maroon}ResNet50\cite{he-cvpr2016-resnet} & 17.6 & 23.5M & 6.6 & 79.8 & 86.4 & 83.1 \\
				\toprule[0.75pt]
			\end{tabular}
		}
\end{table}

\begin{table}[!t]
		\centering
		\caption{Different attacks with HRL. }
		\label{table:abl-variant}
		\small
		\setlength{\tabcolsep}{2.2mm}{
			\begin{tabular}{lcccc}
				\toprule[0.75pt]
				Variants & $\mathcal{J}$  & $\mathcal{F}$  & $\mathcal{J}$\&$\mathcal{F}$ & Gain \\ \midrule[0.5pt]
				\rowcolor{maroon}Origin & 82.2 & 88.6 &  85.4& - \\
				PI\cite{gao-eccv2020-pi-fgsm} & 80.3  & 86.4  & 83.4 & -2.0 \\
				\rowcolor{maroon}PI+HRL  & 78.2  & 85.4  & 81.8 & -3.6 \\
				VMI\cite{wang-cvpr2021-vmi-fgsm} &  80.8 & 86.9  & 83.9 & -1.5  \\
				\rowcolor{maroon}VMI+HRL  & 78.4  & 85.5  & 82.0 & -3.4  \\
				PGD\cite{madry-iclr2018-pgd} & 79.9   &  86.7 & 83.3 &  -2.1 \\
				\rowcolor{maroon}PGD+HRL  &  $\textbf{77.4}$ & $\textbf{85.1}$ & $\textbf{81.3}$ &  $\textbf{-4.1}$ \\
				\toprule[0.75pt]
			\end{tabular}
		}
\end{table}

\begin{table}[!t]
	\centering
	\caption{The hardness loss function in Eq.~(\ref{eq:hardness_loss}).}
	\label{table:abl-hardness-loss}
	\small
	\setlength{\tabcolsep}{3mm}{
		\begin{tabular}{lccc}
			\toprule[0.75pt]
			Loss form & $\mathcal{J}$  & $\mathcal{F}$  & $\mathcal{J}$\&$\mathcal{F}$  \\ \midrule[0.5pt]
			\rowcolor{maroon}MAE:~$||\mathbf{z} -\mathbf{\hat{z}}  ||_1$ & 79.5 & 86.3 & 82.9 \\
			MSE:~$||\mathbf{z} -\mathbf{\hat{z}}  ||_2^2$ & $\textbf{77.4}$  & $\textbf{85.1}$  & $\textbf{81.3}$ \\
			\rowcolor{maroon}~~~CE:~$-\mathbf{z}\log \mathbf{\hat{z}}$  &  78.5 & 85.5 & 82.0  \\
			\toprule[0.75pt]
	\end{tabular}}
\end{table}

\begin{table}[!t]
	\centering
	\caption{Norm of gradient map in Eq.~(\ref{eq:noise_iter}).}
	\label{table:abl-norm}
	\small
	\setlength{\tabcolsep}{3mm}{
		\begin{tabular}{lccc}
			\toprule[0.75pt]
			Norm form & $\mathcal{J}$  & $\mathcal{F}$  & $\mathcal{J}$\&$\mathcal{F}$  \\ \midrule[0.5pt]
			\rowcolor{maroon}$\ell_{\infty}$:~$\epsilon\text{sign}(\mathbf{G}_r)$  & $\textbf{77.4}$  & $\textbf{85.1}$  & $\textbf{81.3}$ \\
			$\ell_{2}$:~$\mathbf{G}_r/ || \mathbf{G}_r ||_2$   & 81.4 & 87.8  & 84.6 \\
			\rowcolor{maroon}$\ell_{1}$:~$\mathbf{G}_r/ || \mathbf{G}_r ||_1$ & 81.6 & 87.9 & 84.8  \\
			\toprule[0.75pt]
	\end{tabular}}
	\vspace{-5mm}
\end{table}

All ablation studies are conducted by attacking the semi-supervised VOS model, \ie, STCN \cite{cheng-nips2021-stcn}, on the DAVIS2017 validation set.

\begin{table*}[!t]
	\centering
	\caption{The transferability of our attacker in terms of $\mathcal{J}$\&$\mathcal{F}$. }  
		\label{table:result-transferability}
	\small
	\setlength{\tabcolsep}{1mm}{
		\begin{tabular}{lc ccc c ccc }
			\toprule[0.75pt]
			\multirow{3}{*}{Attacker} & \multirow{3}{*}{Venue} & \multicolumn{3}{c}{DAVIS2016~\cite{perazzi-cvpr2016-davis2016}} &  & \multicolumn{3}{c}{DAVIS2017~\cite{pont-arxiv2017-davis17}}  \\ 
			\cline{3-5} \cline{7-9} 
			&  &  STCN$\rightarrow$STM & STCN$\rightarrow$HMMN & STCN\cite{cheng-nips2021-stcn} & ~& STCN$\rightarrow$STM & STCN$\rightarrow$HMMN & STCN\cite{cheng-nips2021-stcn} \\ 
			&  & \scriptsize{ICCV'19\cite{oh-iccv2019-stm}} & \scriptsize{ICCV'21\cite{seong-iccv2021-hmmn}}  & \scriptsize{NeurIPS'21} &~ & \scriptsize{ICCV'19\cite{oh-iccv2019-stm}} & \scriptsize{ICCV'21\cite{seong-iccv2021-hmmn}} & \scriptsize{NeurIPS'21}  \\ \midrule[0.5pt]
			\rowcolor{maroon} Origin     & -       &89.3    &90.8   & 91.6   &~    &81.8  &84.7 & 85.4   \\
			PGD\cite{madry-iclr2018-pgd} & ICLR'18 &88.2    &90.4   & 86.9   &~    &81.4  &83.8 & 83.3   \\
			\rowcolor{maroon}            & Gains    &-1.1    &-0.4   & -4.7   &~    &-0.4  &-0.9 & -2.1  \\     
			SegPGD\cite{gu-eccv2022-segpgd} & ECCV'22 &87.9  &89.8 &86.3     &~    &80.8  &83.4 & 83.2 \\
			\rowcolor{maroon}           & Gains     &-1.4    &-1.0   &-5.3    &~    &-1.0  &-1.3 & -2.2 \\     
			\midrule[0.5pt]
			ARA     & Ours & \textbf{87.1} & \textbf{89.2} & \textbf{84.7}   &~    &\textbf{80.3} & \textbf{82.8} &\textbf{81.3}   \\ 
			\rowcolor{maroon}           & Gains     &-2.2    &-1.6   & -6.9   &~    &-1.5  &-1.9 & -4.1  \\     
			\toprule[0.75pt]
		\end{tabular}
	}
\end{table*}

\textbf{HRL Backbone}. To investigate the influence of the backbone on the HRL, several popular deep neural networks are compared in Table~\ref{table:abl-backbone}, including VGG16 \cite{simonyan-iclr2015-vgg}, MobileNetV3-large \cite{howard-iccv2019-mobilenetv3}, InceptionV3 \cite{szegedy-cvpr2016-inceptionv3}, DenseNet121 \cite{huang-cvpr2017-densenet}, and ResNet-18/34/50 \cite{he-cvpr2016-resnet}, which are pre-trained on the ImageNet \cite{deng-cvpr2009-imagenet} database. The last fully-connection layer and pooling layer of these neural networks are removed to derive the feature map of the input frame. Besides the common evaluation metrics, some statistics of these backbones are shown, such as the model parameters in \#Params (Million), the computational complexity in GFLOPs (Giga Floating Point Operations) with a $480\times 854\times 3$ gradient map as the input, and average time of attacking a video in seconds. Among the backbones, ResNet18 achieves the best attack performance with the fastest speed 5.8 s per video and modest parameters. The larger backbones like InceptionV3 and ResNet50 are more difficult to optimize because there are too many parameters to learn, leading to inferior attack results. 

\textbf{Attack Variant}. Table~\ref{table:abl-variant} presents the attack performance of the adversarial example generated by the HRL with several gradient-based attack methods, including PI \cite{gao-eccv2020-pi-fgsm}, VMI \cite{wang-cvpr2021-vmi-fgsm}, and PGD \cite{madry-iclr2018-pgd}. Our attacker takes advantage of unifying both PGD and the HRL, and it degrades the segmentation performance the most by 4.1\% in terms of $\mathcal{J}$\&$\mathcal{F}$, as indicated by the bottom row in Table~\ref{table:abl-variant}. Meanwhile, when using the HRL for PI and VMI, the attack performances are improved by 1.6\% and 1.9\%, respectively, which justifies the good transferability of our HRL.

\textbf{Hardness Loss}. This paper adopts the Mean Squared Error (MSE) as the hardness loss function, and Table~\ref{table:abl-hardness-loss} shows the attack performance of other loss functions, such as Mean Absolute Error (MAE) and CE. Among the three losses, MSE leads to the worst segmentation results when attacking the VOS model, which indicates it is the best choice for computing the hardness loss of the gradient map. Compared with MAE, the MSE function is smoother and easier to optimize. 

\textbf{Norm of Gradient Map}. Table~\ref{table:abl-norm} shows the impacts of different norms applied to the gradient map $\mathbf{G}_r$, including $\ell_{\infty}$-norm, \ie, the $\epsilon\text{sign}(\cdot)$ function in Eq.~(\ref{eq:noise_iter}), $\ell_{2}$-norm, and $\ell_{1}$-norm. It can be seen that the $\ell_{\infty}$-norm has the most powerful attack ability on the segmentation model, and $\ell_{2}$-norm is slightly better than $\ell_{1}$-norm by perturbing the first video frame.

\begin{figure}
	\centering
	\includegraphics[width=0.99\linewidth]{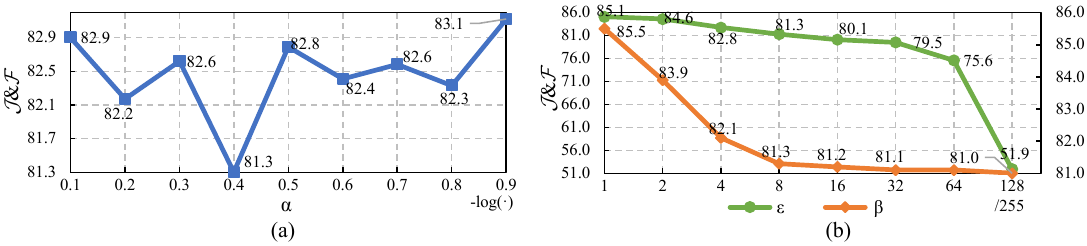}
	\caption{Parameter sensitivity analysis. (a) The label threshold; (b) The perturbation parameter.}
	\label{fig:hyperpara}
	\vspace{-2mm}
\end{figure}

\textbf{Parameter Sensitivity}. When the threshold $\alpha$ of the hardness pseudo-label map varies from $-\log(0.1)$ to $-\log(0.9)$, the attack results are illustrated in Fig.~\ref{fig:hyperpara}(a). When $\alpha$ takes $-\log(0.4)$, the attacker performs the best. Meanwhile, the sensitivity of $\epsilon$ and $\beta$ on the perturbation is investigated by increasing the value from $1/255$ to $128/255$, and the performances are presented in Fig.~\ref{fig:hyperpara}(b). Although the segmentation performance of $\epsilon$ drops quickly after $32/255$, the perturbation becomes more easily visible to human eyes. Thus, following \cite{goodfellow-iclr2015-fgsm}, this paper set $\epsilon$ to $8/255$. Since the performance saturates after $\beta$ equals $16/255$, so the same value of $\epsilon$ is used here.

\begin{figure}[!t]
	\centering
	\includegraphics[width=0.99\linewidth]{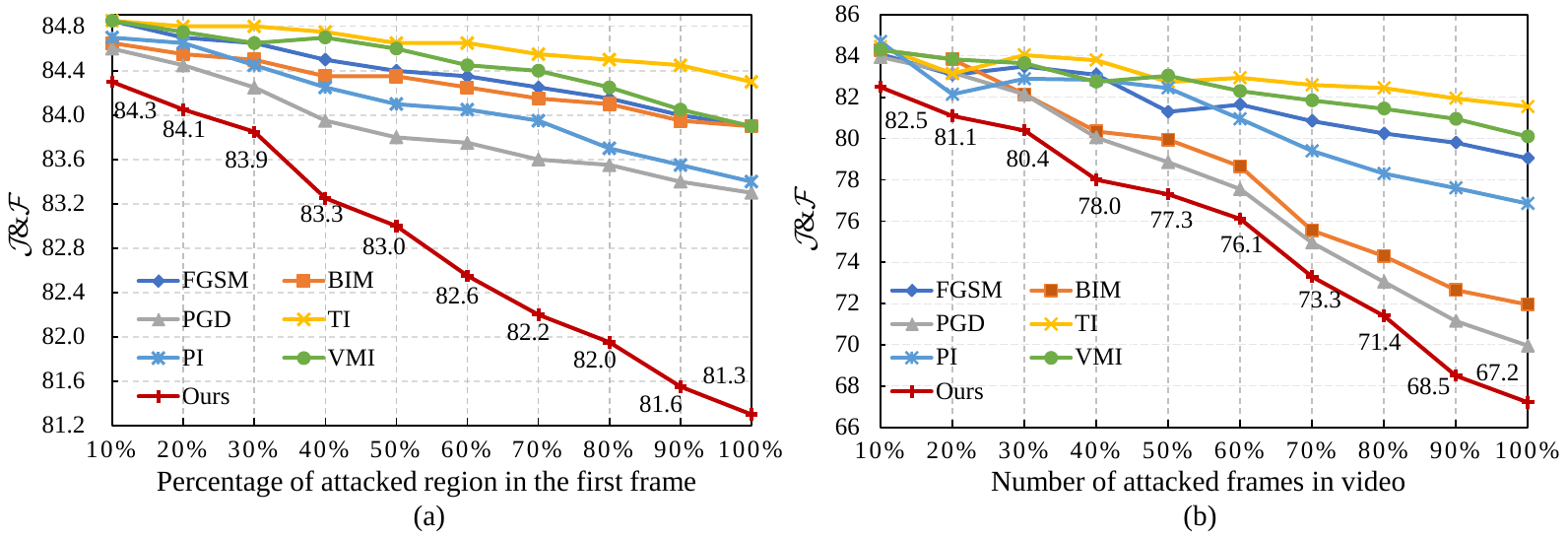}
	\caption{Attacked region and frames. (a) The percentage of the attacked region; (b) The number of attacked frames.}
	\label{fig:attacked-frame-region}
	\vspace{-5mm}
\end{figure}

\textbf{Attacked Region and Frame}. To further explore our attacker, Fig.~\ref{fig:attacked-frame-region} illustrates the VOS performance by adding perturbations to different regions of the first frame and different numbers of frames. It can be seen from Fig.~\ref{fig:attacked-frame-region}(a) that the segmentation performance degrades gradually when the percentage of the attacked region increases from 10\% to 100\%. Compared with other alternatives, such as FGSM \cite{goodfellow-iclr2015-fgsm}, BIM \cite{kurakin-iclr2017-bim}, TI \cite{dong-cvpr2019-ti-fgsm}, and VMI \cite{wang-cvpr2021-vmi-fgsm}, our ARA attacker exhibits more powerful attacking ability as indicated by the much steeper curve. As shown in Fig.~\ref{fig:attacked-frame-region}(b), our attacker greatly degrades the VOS performance with the increasing number of frames attacked, \eg, the $\mathcal{J}$\&$\mathcal{F}$ decreases from 82.5\% to 67.2\% with a margin of 15.3\%. Besides, BIM \cite{kurakin-iclr2017-bim} and PGD \cite{madry-iclr2018-pgd} have stronger attacking power than other alternatives.

\textbf{Attack Transferability}. To explore the transferability of the attack, we choose the STCN \cite{cheng-nips2021-stcn} model as the surrogate, which generates adversarial examples to attack other VOS models, including STM \cite{oh-iccv2019-stm} and HMMN \cite{seong-iccv2021-hmmn}. The transferring attack results are shown in Table~\ref{table:result-transferability}. From the table, we see that the attack transferability seems weak on other segmentation models, which might be the reason that the white-box attack methods usually depend on the model gradients, and the different VOS models propagate distinct gradients, leading to the inferior transferability. Moreover, compared to PGD \cite{madry-iclr2018-pgd} and SegPGD \cite{gu-eccv2022-segpgd}, the model transferability of our ARA attacker is better. This is because our attacker focuses on the hard region area where the foreground and the background are easily confused.

\begin{table*}[!t]
	\centering
	\caption{The VOS segmentation performance with black-box attackers in terms of $\mathcal{J}$\&$\mathcal{F}$. }  
	\label{table:result-davis_blackbox}
	\small
	\setlength{\tabcolsep}{1mm}{
		\begin{tabular}{lc ccc c ccc c ccc}
			\toprule[0.75pt]
			\multirow{3}{*}{Attacker} & \multirow{3}{*}{Venue} & \multicolumn{3}{c}{DAVIS2016~\cite{perazzi-cvpr2016-davis2016}} &  & \multicolumn{3}{c}{DAVIS2017~\cite{pont-arxiv2017-davis17}} & & \multicolumn{3}{c}{YouTube-VOS~\cite{xu-arxiv2018-ytbvos}}  \\ 
			\cline{3-5} \cline{7-9} \cline{11-13}
			&  & \scriptsize{STM\cite{oh-iccv2019-stm}} & \scriptsize{HMMN\cite{seong-iccv2021-hmmn}} & \scriptsize{STCN\cite{cheng-nips2021-stcn}} &  ~& \scriptsize{STM\cite{oh-iccv2019-stm}} & \scriptsize{HMMN\cite{seong-iccv2021-hmmn}} & \scriptsize{STCN\cite{cheng-nips2021-stcn}} &  ~& \scriptsize{STM\cite{oh-iccv2019-stm}} & \scriptsize{HMMN\cite{seong-iccv2021-hmmn}} & \scriptsize{STCN\cite{cheng-nips2021-stcn}} \\ 
			&  & \scriptsize{ICCV'19} & \scriptsize{ICCV'21}  & \scriptsize{NeurIPS'21} &~ & \scriptsize{ICCV'19} & \scriptsize{ICCV'21} & \scriptsize{NeurIPS'21} &~ & \scriptsize{ICCV'19} & \scriptsize{ICCV'21} & \scriptsize{NeurIPS'21}   \\ \midrule[0.5pt]
			\rowcolor{maroon} Origin & - & 89.3 & 90.8 & 91.6  &~ & 81.8 & 84.7 & 85.4  &~ & 79.3 & 82.5 & 84.3 \\
			Random & - & 88.5 & 90.7 & 91.5  & ~& 81.7 & 84.0 & 85.3  &~ & 79.2 & 82.4 & 84.2 \\
			\rowcolor{maroon} OP\cite{su-tevc2019-onepixel} & \scriptsize{TEVC'19} & 87.0 & 88.6 & 89.4  & ~& 80.1 & 80.9 & 82.1  &~ &78.5 & 81.3 & 83.1  \\
			SimBA\cite{guo-icml2019-simba}  & \scriptsize{ICML'19} & 88.1 & 88.4 & 89.6  &~ & 80.1 & 83.4 & 83.1  &~ & 77.8 & 80.2 & 82.6  \\
			\rowcolor{maroon} DE\cite{li-tevc2022-de} & \scriptsize{TEVC'22} & 86.5 & 87.7 & 89.1 & ~& 79.4 & 80.6 & 82.1  &~ & 77.1 & 80.4 & 82.2  \\
			\midrule[0.5pt]
			ARA(Ours) &  & \textbf{84.1} & \textbf{84.3} & \textbf{88.4}  &~ & \textbf{77.5} & \textbf{78.9} & \textbf{81.2}  &~ & \textbf{75.2} & \textbf{77.8} & \textbf{81.0} \\ 
			\toprule[0.75pt]
		\end{tabular}
	}
\end{table*}

\subsection{Black-box Attack Results}
To investigate the performance of our black-box version, this paper compares several SOTA alternatives including OP \cite{su-tevc2019-onepixel}, SimBA \cite{guo-icml2019-simba}, and DE (Differential Evolution) \cite{li-tevc2022-de} to attack the first video frame against the VOS models. Among them, OP generates one-pixel adversarial perturbations based on differential evolution and requires less adversarial information, SimBA randomly samples a vector from a predefined orthonormal basis and either adds to or subtracts it from the target frame, and DE is an approximated gradient sign method that uses differential evolution to solve the black-box adversarial attack problem, by searching the gradient sign rather than the perturbation. The results for three benchmarks are presented in Table~\ref{table:result-davis_blackbox}. 

It can be seen from Table~\ref{table:result-davis_blackbox} that our black-box ARA attacker generates consistently powerful perturbations to the VOS models by attacking only the first video frame, \eg, the VOS performance degrades by at most 6.5\%, 5.8\%, and 4.7\%, on DAVIS2016~\cite{perazzi-cvpr2016-davis2016}, DAVIS2017~\cite{pont-arxiv2017-davis17}, and YouTube-VOS~\cite{xu-arxiv2018-ytbvos}, respectively. Meanwhile, our ARA algorithm leads to larger performance drops than the compared methods, demonstrating its stronger attack power.

\subsection{Defense Results}
To investigate the defense performance of adversarial training, our ARA attacker and the PGD \cite{madry-iclr2018-pgd} attacker are taken as an example to generate attacked frames for training robust semi-supervised VOS models STM \cite{oh-iccv2019-stm} and STCN \cite{cheng-nips2021-stcn}. The results are presented in Table~\ref{table:result-defense}. In detail, perturbations are added to the first frame of the videos from the training sets in DAVIS2017 \cite{pont-arxiv2017-davis17} and YouTube-VOS \cite{xu-arxiv2018-ytbvos}, and the perturbed frames are used to conduct adversarial training for obtaining robust VOS models. Note that the original pre-training with clean videos is completed before the adversarial training with attacked videos, and the perturbation parameters are kept the same as those on the attacker. 

According to Table~\ref{table:result-defense}, the top group shows the VOS performance of the adversarial training or the white-box attack using the PGD and our attacker individually. The first row shows the original segmentation performance as a baseline without any attack or defense. Rows~2 and 3 show that our attacker degrades the performance more significantly than PGD by 2.2\%; Rows~4 and 5 indicate that the VOS model performance is slightly affected by adversarial training of both PGD and our attacker. The bottom group lists the records of the defense results against the attack by the PGD and our attacker. Rows~6 and 7 show that the adversarial training with our method is robust to the VOS model attacked by PGD, and our attacker has higher model robustness than PGD when the video frames are attacked by our ARA method, as indicated by the last two rows. Therefore, the overall defense performance of our ARA method is promising.

\begin{table}[!t]
	\centering
	\caption{The VOS model performance with white-box attackers and adversarial training on the DAVIS2017 \cite{pont-arxiv2017-davis17} val set in terms of $\mathcal{J}$\&$\mathcal{F}$. }
	\label{table:result-defense}
	\small
	\setlength{\tabcolsep}{0.9mm}{
		\begin{tabular}{ccccccrcr}
			\toprule[0.75pt]
			\multicolumn{2}{c}{Attack} &  & \multicolumn{2}{c}{Defense} & STM\cite{oh-iccv2019-stm}  & \multirow{2}{*}{Gains}  & STCN\cite{cheng-nips2021-stcn} & \multirow{2}{*}{Gains} \\ \cline{1-2}  \cline{4-5}
			PGD & Ours & & PGD  & Ours & \scriptsize{ICCV'19} &  & \scriptsize{NeurIPS'21} &    \\  \midrule[0.5pt]
			\rowcolor{maroon} &  &  & &  & 81.8 & -    & 85.4 & - \\
			 	  \checkmark  &  &  & & & 77.2 & -4.6 & 83.3 & -2.1 \\
			 	\rowcolor{maroon}        		& \checkmark & &  &   & 75.0 & -6.8    & 81.3 & -4.1 \\
			     		 &  & & \checkmark  &   & 80.7 & -1.1 & 84.8 & -0.6  \\
			\rowcolor{maroon} &  & & & \checkmark & 81.2 & -0.6 & 84.6 & -0.8 \\ \midrule[0.5pt]
						\checkmark &  & &  \checkmark  &   & 79.0 & -2.8 & 83.4 & -2.0  \\
			  \rowcolor{maroon}               \checkmark &  & &  & \checkmark  & 79.4 & -2.4 & 83.7 & -1.7 \\	
		              	& \checkmark & &  \checkmark  &   & 78.1 & -3.7 & 80.2 & -5.2  \\
			\rowcolor{maroon} & \checkmark & &  & \checkmark & 79.3 & -2.5 & 83.5 & -1.9\\ 		
			\toprule[0.75pt]
		\end{tabular}
	}
\end{table}

\begin{figure}
\centering
	\includegraphics[width=\linewidth]{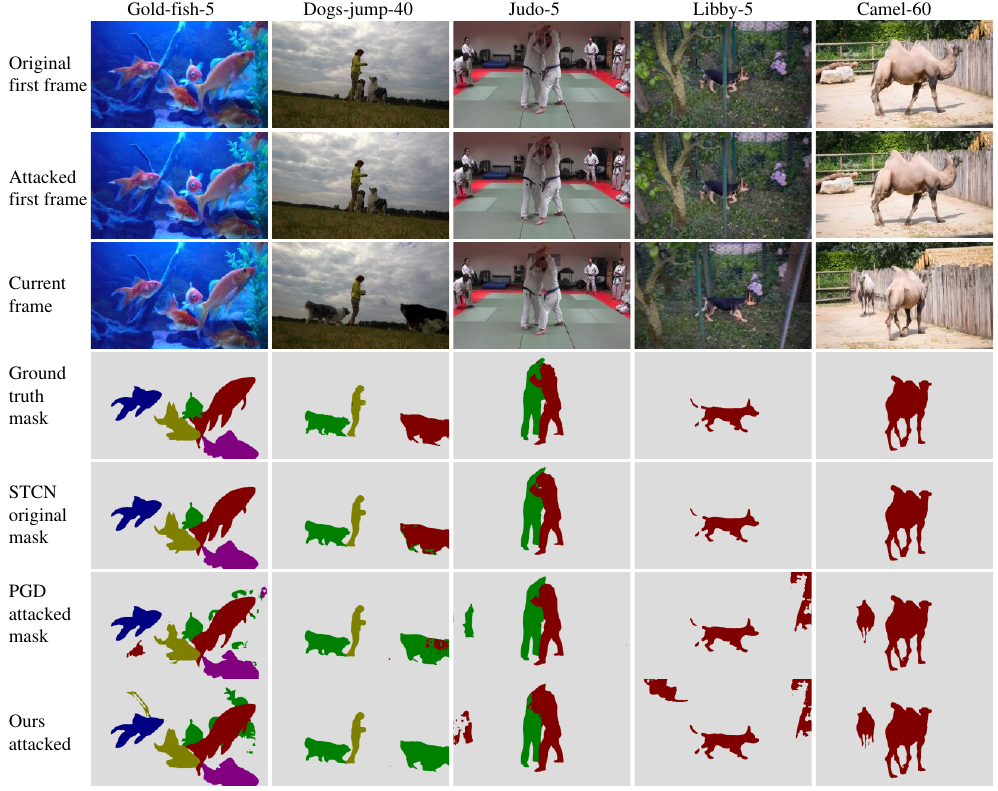}
	\caption{The visualization of the attack results on the STCN \cite{cheng-nips2021-stcn} model.}
	\label{fig:vis}
\end{figure}

\subsection{Qualitative Results}
To illustrate the attack performance, several videos were randomly chosen from DAVIS2017 \cite{pont-arxiv2017-davis17}, and the first frame of each video is attacked against the VOS model, \ie, STCN \cite{cheng-nips2021-stcn}, which adopts ResNet as the backbone. As illustrated in Fig.~\ref{fig:vis} where each color indicates one object, it is almost impossible to perceive the perturbation added to the adversarial example (Row~2) by human eyes, indicating the good safety of our attacker. 

For the test examples in Row~3, the VOS model obtains satisfactory segmentation results (Row~4), while the PGD \cite{madry-iclr2018-pgd} attacker and our ARA attacker (including white-box and black-box) successfully fool the model to make incorrect pixel predictions indicated by the red area of Rows~5 to 7. For the PGD attacker, the proportion of its prediction error pixels is less than that of our attacker, demonstrating that the developed ARA attacker produces adversarial examples with a stronger attacking power. Also, for our ARA attacker, the power of the white-box attacker is better than that of the black-box one, \ie, the red error area is larger. This is because the white-box setting can employ the gradients during the optimization, but the gradients are unavailable for the black-box setting. Moreover, our method has a satisfactory defense effect against the ARA attacker, as the red error area is greatly reduced compared to those results in the above rows. 

\section{Conclusion}
\label{conclusion}

This work explores the effects of adversarial attacks on video object segmentation. An adversarial region attacker is developed to generate adversarial examples by adding almost human-imperceptible perturbations to the first frame of the video. Meanwhile, to improve the attack power of the adversarial example, a hard region learner is introduced to derive the hardness map by using the gradients derived from the model back-propagation mechanism. This makes the perturbation emphasize the pixel areas where the foreground and the background are easily confused. Moreover, the iterative strategy is adopted to update the perturbation, thus improving the attack ability of the adversarial example. Finally, extensive experiments are conducted on three benchmarks to verify the superiority of our attacker to other adversarial attack methods on several state-of-the-art VOS models. In the future, we will investigate attacking the frames with large uncertainty to enhance the attack power.



\ifCLASSOPTIONcaptionsoff
  \newpage
\fi

\end{document}